\documentclass[sigconf]{acmart}

\AtBeginDocument{%
  \providecommand\BibTeX{{%
    \normalfont B\kern-0.5em{\scshape i\kern-0.25em b}\kern-0.8em\TeX}}}

\copyrightyear{2024}
\acmYear{2024}
\setcopyright{rightsretained}
\acmConference[KDD '24]{Proceedings of the 30th ACM SIGKDD Conference on Knowledge Discovery and Data Mining}{August 25--29, 2024}{Barcelona, Spain}
\acmBooktitle{Proceedings of the 30th ACM SIGKDD Conference on Knowledge Discovery and Data Mining (KDD '24), August 25--29, 2024, Barcelona, Spain}\acmDOI{10.1145/XXXXXXX.XXXXXXX}
\acmISBN{979-8-4007-0490-1/24/08}

\begin{document}

\title{EmoLLMs: A Series of Emotional Large Language Models and Annotation Tools for Comprehensive Affective Analysis}

\author{Zhiwei Liu}
\authornote{Corresponding author}
\affiliation{%
  \institution{The University of Manchester}
  \city{Manchester}
  \country{United Kingdom}
  \postcode{M1 7DN}
}
\email{zhiwei.liu-2@postgrad.manchester.ac.uk}

\author{Kailai Yang}
\affiliation{%
  \institution{The University of Manchester}
  \city{Manchester}
  \country{United Kingdom}
  \postcode{M1 7DN}
}
\email{kailai.yang@postgrad.manchester.ac.uk}

\author{Qianqian Xie}
\authornote{Qianqian is now affiliated with Yale University. The work was done while she was at The University of Manchester.}
\affiliation{%
  \institution{The University of Manchester}
  \city{Manchester}
  \country{United Kingdom}
  \postcode{M1 7DN}
}
\email{xqq.sincere@gmail.com}

\author{Tianlin Zhang}
\affiliation{%
  \institution{The University of Manchester}
  \city{Manchester}
  \country{United Kingdom}
  \postcode{M1 7DN}
}
\email{zhangtianlin668@gmail.com}

\author{Sophia Ananiadou}
\affiliation{%
  \institution{The University of Manchester}
  \city{Manchester}
  \country{United Kingdom}
  \postcode{M1 7DN}
}

\affiliation{%
  \institution{Artificial Intelligence Research Center}
  \city{Tokyo}
  \country{Japan}
}

\email{sophia.ananiadou@manchester.ac.uk}

\renewcommand{\shortauthors}{Zhiwei Liu, Kailai Yang, Qianqian Xie, Tianlin Zhang, \& Sophia Ananiadou}

\begin{abstract}
  Sentiment analysis and emotion detection are important research topics in natural language processing (NLP) and benefit many downstream tasks. With the widespread application of large language models (LLMs), researchers have started exploring the application of LLMs based on instruction-tuning in the field of sentiment analysis. However, these models only focus on single aspects of affective classification tasks (e.g. sentimental polarity or categorical emotions), and overlook the regression tasks (e.g. sentiment strength or emotion intensity), which leads to poor performance in downstream tasks. The main reason is the lack of comprehensive affective instruction tuning datasets and evaluation benchmarks, which cover various affective classification and regression tasks. Moreover, although emotional information is useful for downstream tasks, existing downstream datasets lack high-quality and comprehensive affective annotations. In this paper, we propose EmoLLMs, the first series of open-sourced instruction-following LLMs for comprehensive affective analysis based on fine-tuning various LLMs with instruction data, the first multi-task affective analysis instruction dataset (AAID) with 234K data samples based on 3 classification tasks and 2 regression tasks to support LLM instruction tuning, and a comprehensive affective evaluation benchmark (AEB) with 8 regression tasks and 6 classification tasks from various sources and domains to test the generalization ability of LLMs.  We propose a series of EmoLLMs by fine-tuning LLMs with AAID to solve various affective instruction tasks. We compare our models with a variety of LLMs and sentiment analysis tools on AEB, where our models outperform all other open-sourced LLMs and sentiment analysis tools, and surpass ChatGPT and GPT-4 in most tasks, which shows that the series of EmoLLMs achieve the ChatGPT-level and GPT-4-level generalization capabilities on affective analysis tasks, and demonstrates our models can be used as affective annotation tools. This project is available at https://github.com/lzw108/EmoLLMs/.
\end{abstract}




\maketitle

\section{Introduction}
Emotions and sentiments play a crucial role in shaping our lives. Our words and actions serve as indicators of our emotional states \cite{liu2020sentiment1}. Leveraging natural language processing (NLP) techniques such as Emotion Detection (ED) and Sentiment Analysis (SA), we can delve into the analysis of human interactions, enabling us to comprehend people's emotional responses toward particular subjects \cite{8004002Hakak2017}. Specifically, SA tasks typically involve predicting the polarity (usually positive, negative, or neutral), along with the strength of this tone \cite{thelwall2010sentiment1}, and emotion detection tasks often involve classifying data into fine-grained emotion categories (e.g. Ekman \cite{ekman1992argument2}, Plutchik \cite{plutchik1980general3}) or predicting the intensity of emotions \cite{xie2018novel2}. These affective information are proven as useful features for many downstream tasks, including mental health analysis~\cite{zhang2023emotion1}, misinformation detection \cite{liu2023emotion2}, and empathetic dialogue systems \cite{ma2020survey}.

Pre-trained language models (PLMs) such as BERT~\cite{devlin2018bert} and RoBERTa \cite{liu2019roberta}, have shown excellent performance in classification tasks. Many studies have applied them to sentiment analysis or emotion detection tasks \cite{bello2023bert1,liao2021improved2,yin2020sentibert1}. However, these PLMs are limited by the scale of model parameters and the training corpus, resulting in a lack of comprehensive understanding and generalization ability for complex tasks \cite{zhang2023enhancing1}, which restricts the effectiveness of affective analysis, especially in affective regression tasks \cite{zhang2020affect2}. 
Compared to PLMs, LLMs have the characteristic of having an enormous parameter size, typically reaching hundreds of billions or more, which gives them stronger generalization abilities in downstream tasks and enables them to handle tasks with intricate complexities \cite{zhang2023dialoguellm2}. Many researchers have started exploring the application of LLMs in the field of sentiment analysis, which achieved excellent performance by fine-tuning open-source LLMs on sentiment analysis tasks \cite{zhang2023enhancing1,zhang2023dialoguellm2,lei2023instructerc4}. However, these studies only focus on sentiment/emotion classification tasks and overlook regression tasks (e.g. sentiment strength, emotion intensity), which provide more fine-grained affective features \cite{akhtar2020intense3} and are proven useful in many scenarios \cite{zhang2021mining3,qureshi2020improving,consoli2022fine3}. The major reason is the lack of a comprehensive instruction-based sentiment analysis dataset and evaluation benchmark.

Moreover, though emotional information is proven useful for downstream tasks, existing downstream datasets lack emotion-related resources such as sentiment/emotion labels. Therefore, many works use affective analysis tools (e.g. VADER \cite{hutto2014vader}, TextBlob\footnote{https://textblob.readthedocs.io/}) to provide sentiment annotations. For example, in \cite{govindasamy2021depression}, the authors utilized the TextBlob library to calculate sentiment scores and fed them into a depression detection classifier. 
Additionally, some studies employ transfer learning methods, by applying models trained on other sentiment analysis or emotion-labeled datasets to automatically annotate the emotions expressed in downstream task datasets \cite{chan2023state3,dong2022sentiment1hh}. However, these tools or methods can only annotate one aspect of sentiment analysis tasks, resulting in limited coverage of emotional features.

To address the above issues, we propose a suite of LLMs, instruction-tuning datasets, and an evaluation benchmark for multi-task affective analysis. 
We first construct the multi-task affective analysis instruction dataset (AAID) with 234K data samples to support LLM instruction tuning, which is based on SemEval-2018 Task1: Affect in Tweet \cite{mohammad2018semeval1,mohammadkiritchenko2018understanding2}, including five tasks: emotion intensity regression, ordinal classification of emotion intensity, sentiment strength regression, sentiment classification, and multi-label emotion classification. Based on the AAID dataset, we propose a series of emotional large language models (EmoLLMs), the first open-sourced instruction-following LLMs for comprehensive affective analysis, by performing multi-task instruction tuning on LLMs. To evaluate the performance and generalizability of EmoLLMs, we also construct an affective evaluation benchmark (AEB) based on 14 affective analysis datasets collected from various platforms and sources, which include 8 regression tasks and 6 classification tasks.

Based on AEB, we evaluate EmoLLMs, a variety of open-sourced LLMs, close-sourced LLMs (i.e. ChatGPT and GPT-4), and several sentiment analysis tools. The experimental results indicate that the series of EmoLLMs overtake all other open-sourced LLMs, and sentiment analysis tools, and exceed ChatGPT and GPT-4 in 7 regression tasks and 4 classification tasks. These results demonstrate that EmoLLMs achieve a comparable capability with ChatGPT and GPT-4 in most affective analysis tasks. EmoLLMs can serve as comprehensive affective annotation tools for annotating data from different platforms and sources. 

Our main contributions are as follows: 
\begin{itemize}

\item We build AAID, the first multi-task affective analysis instruction tuning data, and AEB, the first affective generalization testing instruction benchmark.

\item We introduce a series of EmoLLMs, the first open-source instruction following LLMs for comprehensive affective analysis.

\item We compare EmoLLMs with other LLMs on AEB. Additionally, we conduct a comprehensive analysis of the affective analysis capabilities of ChatGPT and GPT-4. Our models achieve SOTA performance on the AEB dataset compared to other open-sourced LLMs and present ChatGPT-level and GPT-4-level generalization capabilities, establishing their potential as effective tools for affective annotation.

\end{itemize}

The structure of this paper is as follows: Section \ref{sec:relatedwork} introduces the related work about sentiment analysis models and open-sourced LLMs. Section \ref{sec:methods} introduces the proposed method. Specifically, Section \ref{definition} introduce the task definition. Section \ref{sec:dataconstruct} and Section \ref{sec:AEB} present the construction process of AAID and AEB respectively. Section \ref{sec:emollama} introduces the training process of EmoLLMs. Section \ref{sec:results} presents the experiment results on AEB and analyses the performance of each model. Section \ref{sec:conclusion} concludes this paper by summarizing our findings. Section \ref{sec:discussions} discusses the real-world applications of EmoLLMs, limitations, and future work.

\section{Related Work \label{sec:relatedwork}}

\subsection{Affective Analysis Model}

There have been various affective analysis tools proposed, such as VADER \cite{hutto2014vader}, and TextBlob. Although these tools are convenient to use, their effectiveness in sentiment analysis is not ideal \cite{he2022they66}. In recent years, many studies have focused on fine-tuning PLMs to enhance their capabilities in the field of sentiment analysis. Bello et al. \cite{bello2023bert1} combine BERT with other deep learning models (e.g. CNN, RNN, LSTM) to improve the ability of the model in short and simple text sentiment analysis. Liao et al. \cite{liao2021improved2} propose a multi-task model based on RoBERTa for aspect-category sentiment analysis. Yin et al. \cite{yin2020sentibert1} propose the SentiBERT model, which focuses on the field of sentiment analysis. SentiBERT integrates a recursive constituency tree based on BERT to better capture compositional sentiment semantics. Recently, numerous studies have embarked on investigating the utilization of LLMs in sentiment analysis, resulting in remarkable performance gains in sentiment analysis tasks. Zhang et al. \cite{zhang2023enhancing1} propose a retrieval-augmented LLM for financial sentiment analysis, which utilizes additional background information from external sources and outperforms LLM baselines by 15\% and 48\%.  Similarly, Lei et al. \cite{lei2023instructerc4} also use a simple yet effective retrieval module to enhance the emotion recognition capability of LLM in dialogue. Zhang et al. \cite{zhang2023dialoguellm2} develop a context and emotion knowledge-tuned LLM, namely DialogueLLM, obtained by fine-tuning LLM with multimodal (i.e., texts and videos) emotional dialogues, which achieved SOTA results on three emotion recognition in conversations (ERC) datasets. However, these PLMs and LLMs only focus on individual aspects of affective analysis, lacking the ability to predict sentiment strength and emotion intensity.

\subsection{Open Sourced Large Language Models}

Although ChatGPT and GPT-4 have shown excellent performance in various fields, their closed-source availability affects the progress of scientific research. Therefore, numerous studies are dedicated to democratizing LLMs, such as the LLaMA series \cite{touvron2023llama1,touvron2023llama2}, OPT series \cite{zhang2022opt6}, BLOOM series \cite{workshop2022bloom7}, and Falcon \cite{penedo2023refinedweb}. Based on the open-source LLMs, many efforts have been made to develop models with instruction-following capabilities like ChatGPT by training on extensive instruction-tuning datasets (e.g. Alpaca\footnote{https://crfm.stanford.edu/2023/03/13/alpaca.html} and the Vicuna\footnote{https://lmsys.org/blog/2023-03-30-vicuna/}). Recently, there has been a lot of domain-specific work aimed at improving the performance of LLM in specific domains by training on domain-specific instruction datasets. Such as FinMA \cite{xie2023pixiu2} in the finance domain, MentalLLaMA \cite{yang2023mentalllama3} in the mental health domain, TimeLlaMA \cite{yuan2023back1} used for temporal reasoning, and ExTES-LLaMA \cite{zheng2023building} in emotional support chatbots. Our work is the first open-sourced LLM series for comprehensive multitask affective analysis.

\section{Methods \label{sec:methods}}

The goal of this work is to evaluate and enhance the comprehensive and complex affective analysis capabilities of LLMs. To achieve this objective, we build the first affective analysis instruction dataset (AAID) to support LLMs tuning for comprehensive affective analysis tasks. We propose EmoLLMs, a series of emotional LLMs by fine-tuning LLMs based on AAID. Furthermore, we construct a comprehensive affective evaluation benchmark to test the generalization ability of LLMs.

\subsection{Task Definition \label{definition}}

Similar to \cite{yang2023mentalllama3} in handling mental health analysis tasks, we also approach affective analysis as a generative task, where a generative model (i.e., an autoregressive language model $P_{\phi}(y|x)$ parameterized by pre-trained weights $\phi$) is employed as the foundation, which is unlike previous discriminative and regression models. This model is capable of simultaneously addressing $N$ affective analysis tasks, such as sentiment polarity and strength prediction, emotion classification and intensity prediction. Each task t is represented by a subset of training context-target pairs: $D_t={(q_i^t,r_i^t)}_{i={1,2,...N_t}}$, where $q$ is a token sequence containing the task description, target text, and query, and $r$ is another sequence containing the query answer (i.e., classification result or regression result). All subsets are combined into a training dataset: $D$. The model is optimized based on this merged data, aiming to maximize the conditional language modeling objective to enhance the accuracy of predictions.

\subsection{Instruction Tuning Data Building \label{sec:dataconstruct}}

We build the instruction dataset based on the SemEval-2018 Task 1: Affect in Tweets, which includes a series of highly annotated sentiment analysis subtasks \cite{mohammad2018semeval1,mohammadkiritchenko2018understanding2}.

\subsubsection{Raw Data}

SemEval 2018 Task1 contains five subtasks: 1. emotion intensity regression (\textbf{EI-reg}), 2. ordinal classification of emotion intensity (\textbf{EI-oc}), 3. valence (sentiment) regression (\textbf{V-reg}), 4. ordinal classification of valence (sentiment) (\textbf{V-oc}), and 5. emotion classification (\textbf{E-c}). 

\textbf{EI-reg}: Given a tweet and an emotion E (anger, fear, joy, sadness), determine the intensity of E that best represents the mental state of the tweeter—a real-valued score between 0 (least E) and 1 (most E);

\textbf{EI-oc}: Given a tweet and an emotion E (anger, fear, joy, sadness), classify the tweet into one of four ordinal classes (0: no E can be inferred. 1: low amount of E can be inferred. 2: moderate amount of E can be inferred. 3: high amount of E can be inferred) of intensity of E that best represents the mental state of the tweeter;

\textbf{V-reg}: Given a tweet, determine the intensity of sentiment or valence (V) that best represents the mental state of the tweeter—a real-valued score between 0 (most negative) and 1 (most positive);

\textbf{V-oc}: Given a tweet, classify it into one of seven ordinal classes (from -3: very negative to 3: very positive), corresponding to various levels of positive and negative sentiment intensity, that best represents the mental state of the tweeter;

\textbf{E-c}: Given a tweet, classify it as ‘neutral or no emotion’ or as one, or more, of eleven given emotions (anger, anticipation, disgust, fear, joy, love, optimism, pessimism, sadness, surprise, trust) that best represent the mental state of the tweeter.

\begin{table}[]
\caption{Statistics of the data. 'Raw' denotes the raw data from SemEval-2018 Task 1: Affect in Tweets. 'Instruction' denotes the converted instruction data based on raw data.}
\label{tab:statisticsofdata} 
\begin{tabular}{p{2cm}p{2cm}p{2cm}p{1cm}}
\hline
Task & Raw (Train/Dev) & Instruction (Train/Dev) & Source  \\
\hline
EI-reg, EI-oc                &                 &                         &         \\
\multicolumn{1}{c}{anger}    & 1701/388        & 17010/3880              & Twitter \\
\multicolumn{1}{c}{fear}     & 2252/389        & 22520/3890              & Twitter \\
\multicolumn{1}{c}{joy}      & 1616/290        & 16160/2900              & Twitter \\
\multicolumn{1}{c}{sadness}  & 1533/397        & 15330/3970              & Twitter \\
V-reg, V-oc                  & 1181/449        & 11810/4490              & Twitter \\
E-c                          & 6838/886        & 68380/8860              & Twitter   \\
\hline
\end{tabular}
\end{table}

\subsubsection{AAID: Affective Analysis Instruction Dataset}

\begin{figure*}[]
\centering
\includegraphics[width=2\columnwidth]{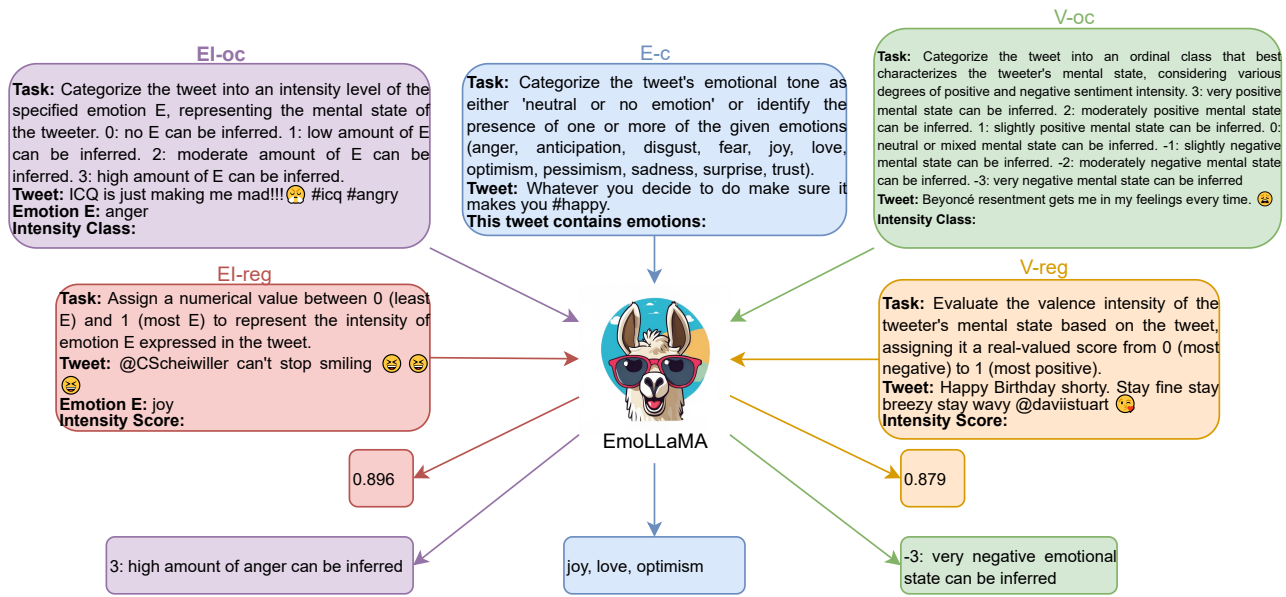}
\caption{An overview of multi-task instruction tuning of EmoLLaMA for multiple affective analysis tasks.}
\label{fig:examplesofemollama}
\end{figure*}

We construct the instruction dataset based on the raw data. Due to the limited quantity of the original dataset, we utilize 10 different task instructions for each task to augment the training set and validation set. The data statistics are presented in Table \ref{tab:statisticsofdata}. Specifically, we build instruction-tuning samples based on some templates. Table \ref{tab:templates} describes the specific instruction templates for each task, and Figure \ref{fig:examplesofemollama} provides corresponding examples (Taking EmoLLaMA as the example and each task selects one [\textit{task prompt}] as an example). [\textit{task prompt}] describes the instructions for each specific task. The word 'Tweet' can be adjusted based on the actual task. The [\textit{input text}] refers to the content of the raw data. The final [\textit{output}] should be adjusted based on the specific task to provide sentiment classification, sentiment strength, emotion classification, or emotion intensity.

\begin{table}[]
\caption{Templates for constructing prompts for instruction dataset. [\textit{task prompt}] denotes task instruction, [\textit{input text}] is from the raw data, [\textit{emotion}] can be anger, fear, joy, sadness, [\textit{output}] is the output from LLM.}
\label{tab:templates} 
\begin{tabular}{p{0.9cm}p{6.7cm}}
\hline
Task   & Prompt Template                                                                                        \\ \hline
EI-reg & Task: {[}\textit{task prompt}{]} Tweet: {[}\textit{input text}{]} Emotion E: {[}\textit{emotion}{]} Intensity score: {[}\textit{output}{]} \\
EI-oc  & Task: {[}\textit{task prompt}{]} Tweet: {[}\textit{input text}{]} Emotion E: {[}\textit{emotion}{]} Intensity class: {[}\textit{output}{]} \\
V-reg  & Task: {[}\textit{task prompt}{]} Tweet: {[}\textit{input text}{]} Intensity score: {[}\textit{output}{]}                          \\
V-oc   & Task: {[}\textit{task prompt}{]} Tweet: {[}\textit{input text}{]} Intensity class: {[}\textit{output}{]}                          \\
E-c    & Task: {[}\textit{task prompt}{]} Tweet: {[}\textit{input text}{]} This tweet contains emotions: {[}\textit{output}{]}             \\ \hline
\end{tabular}
\end{table}

\subsection{AEB: Affective Evaluation Benchmark Building \label{sec:AEB}}

\begin{table*}[]
\footnotesize
\caption{Statistics of AEB. 'R' denotes the regression task, followed by the intensity range. 'SC' denotes the sentiment classification task and 'EC' denotes the emotion classification task, followed by the number of categories.}
\label{tab:statisticsoggtd} 
\begin{tabular}{lccllccllccl}
\hline
Dataset & Size & Type   & Source  & Dataset  & Size & Type     & Source            & Dataset   & Size                 & Type                 & Source        \\ \hline
EI-reg                          & 4068 & R[0,1] & Twitter & V-Amazon & 1000 & R [-4,4] & Amazon            & SST       & 2210                 & R [0,1]              & Movie reviews \\
EI-oc                           & 4068 & EC(4)  & Twitter & V-Movies & 1000 & R [-4,4] & Movies reviews    & SST-5     & 2210                 & SC (5)               & Movie reviews \\
V-reg                           & 937  & R[0,1] & Twitter & V-NYT    & 1000 & R [-4,4] & New York Times    & TDT       & 692                  & SC (3)               & Twitter       \\
V-oc                            & 937  & SC(7)  & Twitter & V-Tweet  & 1000 & R [-4,4] & Twitter           & GoEmotion & 5427                 & EC (7)               & Reddit        \\
E-c                             & 3259 & EC(11) & Twitter & EmoBank  & 1000 & R [1,5]  & News, blogs. etc. &           & \multicolumn{1}{l}{} & \multicolumn{1}{l}{} &               \\ \hline
\end{tabular}
\end{table*}

\begin{table*}[]
\scriptsize
\caption{The task prompt example for each dataset in AEB.}
\label{tab:taskprompt} 
\begin{tabular}{p{1cm}p{16cm}}
\hline
Dataset & Task prompt\\ 
\hline
EI-reg                          & Assign a numerical   value between 0 (least E) and 1 (most E) to represent the intensity of   emotion E expressed in the tweet.                                                                                                                                                                                                                                                                                                                                                                                                                           \\
EI-oc                           & Categorize the tweet   into an intensity level of the specified emotion E, representing the mental   state of the tweeter. 0: no E can be inferred. 1: low amount of E can be   inferred. 2: moderate amount of E can be inferred. 3: high amount of E can be   inferred.                                                                                                                                                                                                                                                                                 \\
V-reg                           & Evaluate the valence   intensity of the tweeter's mental state based on the tweet, assigning it a   real-valued score from 0 (most negative) to 1 (most positive).                                                                                                                                                                                                                                                                                                                                                                                        \\
V-oc                            & Categorize the tweet   into an ordinal class that best characterizes the tweeter's mental state,   considering various degrees of positive and negative sentiment intensity. 3:   very positive mental state can be inferred. 2: moderately positive mental   state can be inferred. 1: slightly positive mental state can be inferred. 0:   neutral or mixed mental state can be inferred. -1: slightly negative mental   state can be inferred. -2: moderately negative mental state can be inferred.   -3: very negative mental state can be inferred. \\
E-c                             & Categorize the   tweet's emotional tone as either 'neutral or no emotion' or identify the   presence of one or more of the given emotions (anger, anticipation, disgust,   fear, joy, love, optimism, pessimism, sadness, surprise, trust).                                                                                                                                                                                                                                                                                                               \\

V-A, V-M, V-NYT, V-T            & Calculate the   sentiment intensity or valence score of the text, which should be a real   number between -4 (extremely negative) and 4 (extremely positive).                                                                                                                                                                                                                                                                                                                                                                                             \\
SST                             & Calculate the   sentiment score of the text, which should be a real number between 0   (extremely negative) and 1 (extremely positive).                                                                                                                                                                                                                                                                                                                                                                                                                   \\
Emobank                         & Determine the   valence/arousal/dominance intensity of the writer's mental state on a scale   of 1 (most negative) to 5 (most positive).                                                                                                                                                                                                                                                                                                                                                                                                                  \\
GoEmotion                       & Categorize the   text's emotional expression, classifying it as either 'neutral' or as one or   more of the specified emotions (anger, disgust, fear, joy, sadness, surprise)   that reflect the writer's state of mind.                                                                                                                                                                                                                                                                                                                                  \\
SST5                            & Classify the text   into one of five classes of sentiment that best represents the mental state   of the text. 0: very negative, 1: negative, 2: neutral, 3: positive, 4: very   positive.                                                                                                                                                                                                                                                                                                                                                                \\
TDT                             & Classify the text   into one of three classes of sentiment that best represents the mental state   of the text. -1: negative, 0: neutral, 1: positive.                                                                                                                                                                                                                                                                                                                                                                                                    \\ \hline
\end{tabular}
\end{table*}

We first collect the test data from SemEval-2018 Task 1: Affect in Tweets. To test the robustness of our model, a random instruction from the ten instructions used in train augment is selected for each instance in the test set. We also collect additional sentiment analysis or emotion detection datasets from various sources and domains to test the generalizability of our model. We construct the AEB following the template format provided in Table \ref{tab:templates}. Table \ref{tab:taskprompt} shows the task prompt example for each dataset. Except for the four datasets from VADER, all other datasets utilize the original test dataset. Table \ref{tab:statisticsoggtd} shows the statistic details.

\textbf{Datasets used in Valence  Aware  Dictionary for sEntiment Reasoning (VADER)} \cite{hutto2014vader1}: There are four datasets from different social media platforms with sentiment intensity (Valence) scores within [-4,4]: \textbf{V-Amazon} (Amazon reviews snippets), \textbf{V-Movies} (Movies reviews snippets, collected from rotten.tomatoes.com), \textbf{V-NYT} (New York Times editorial snippets), \textbf{V-Tweet} (Tweets). We randomly sampled 1000 instances from each dataset for generalizability testing.

\textbf{EmoBank} \cite{buechel2022emobank1,buechel2017readers2}: This dataset was collected from News, blogs, fictions, letters etc. and contains three dimensions, which were manually annotated with emotion according to the psychological Valence-Arousal-Dominance scheme with scores within [1,5].

\textbf{Stanford Sentiment Treebank (SST)} \cite{socher2013recursive1}: It is collected from movie reviews, which is the first corpus with fully labeled parse trees, allowing for a comprehensive analysis of the compositionality of sentiment in language. In SST\footnote{https://huggingface.co/datasets/sst}, each sentence is assigned a floating-point label that indicates the degree of positive sentiment, ranging from 0.0 to 1.0. while in SST5\footnote{https://huggingface.co/datasets/SetFit/sst5}, each sentence is annotated with five labels: very positive, positive, neutral, negative, very negative.

\textbf{Target Dependent Twitter Sentiment Classification (TDT)} \cite{dong2014adaptive2}:  It is a Twitter sentiment classification dataset collected from post comments for the celebrities, products, and companies, which is annotated manually with three labels (negative, neutral, positive). To facilitate our generalizability testing, we restored the entities that were masked in the original data, creating a standard sentence-level sentiment analysis dataset.

\textbf{GoEmotion} \cite{demszky2020goemotions1}: It is a multi-label classification dataset collected from Reddit comments, which consists of 28 emotion labels, including the neutral. However, the original dataset with 28 emotion labels is imbalanced. To mitigate this issue, we select the "Ekman" option from the dataset provided by the authors, which consists of 7 emotion labels, including the neutral.

Since the first five datasets are collected from the same sources as the AAID, the remaining data comes from different platforms and sources, we divide AEB into two parts for comparison. The former is referred to as AEB-1, used to test the training effectiveness of the models. The latter is called AEB-2, which is suitable for testing the generalization ability of models.

\subsection{EmoLLMs \label{sec:emollama}}

We build EmoLLMs by fine-tuning various LLMs based on AAID. We train three EmoLLaMA models based on LLaMA2 \cite{touvron2023llama2}: EmoLLaMA-7B, EmoLLaMA-chat-7B, EmoLLaMA-chat-13B by fine-tuning LLaMA2-7B, LLaMA2-chat-7B, LLaMA2-chat-13B, where LLaMA2-chat-7B and LLaMA2-chat-13B are the first open-source LLMs tuned with reinforcement
learning from human feedback (RLHF) \cite{stiennon2020learning3}. We also train EmoOPT, and EmoBLOOM based on OPT-13B \cite{zhang2022opt6} and BLOOM-7B \cite{workshop2022bloom7}) respectively. All models are trained for three epochs based on AdamW optimizer \cite{loshchilov2017decoupled4}, utilizing early stopping techniques \cite{dodge2020fine1} to prevent overfitting, and leveraging DeepSpeed \cite{rasley2020deepspeed1,yao2023deepspeed2} to reduce memory usage. We set the batch size to 256. The initial learning rate is set to 1e-6 with a warm-up ratio of 5\%, and the maximum model input length is set to 2048. All models are trained on two Nvidia Tesla A100 GPUs, each with 80GB of memory.

\section{Evaluation \label{sec:results}}

\subsection{Base Models}

\textbf{PLMs:} Sentiment analysis and emotion detection are typically regarded as classification tasks, while intensity prediction is considered a regression task. We select some commonly used PLMs as baseline models, which can only fine-tuned on a single task, including BERT, RoBERTa, and one domain-specific pre-trained model (i.e. SentiBERT \cite{yin2020sentibert1}). We add a fully connected neural layer to each model, which is used for classification or regression. For EI-reg and V-reg tasks, we utilize the mean squared error (MSE) loss function. For EI-oc and V-oc tasks, we use cross-entropy loss. For multi-label task E-c, we adopt binary cross-entropy with logits loss.

\textbf{Zero-shot/few-shot methods (LLMs without fine-tuning):} With the emergence of LLMs, zero-shot and few-shot learning have become effective approaches for solving numerous tasks. We select Falcon-7b-instruct \cite{penedo2023refinedweb}, Vicuna-13b-v1.5\footnote{https://huggingface.co/lmsys/vicuna-13b-v1.5}, LLaMA2-chat-7B and LLaMA2-chat-13B to perform zero-shot prompting on the instruction dataset. In addition, we employ zero-shot and few-shot prompting methodologies with the closed-source LLM ChatGPT (gpt-3.5-turbo) and GPT-4 (gpt-4-1106-preview). We select at least one piece of data for each emotion category or label category to serve as few-shot prompts.

\textbf{Emotion-based instruction-tuning methods:} In addition to the EmoLLMs series models, we also fine-tuned BART \cite{lewis2019bart4}, T5 \cite{raffel2020exploring5}  using the same instructional dataset as baseline models to further evaluate the effectiveness of our models.

\subsection{Evaluation Methods}

For AEB-1, we use the official evaluation metric\footnote{https://competitions.codalab.org/competitions/17751}, Pearson correlation coefficient (pcc), as the evaluation metric for EI-reg, EI-oc, V-reg, and V-oc and use accuracy, micro-F1 (mi-F1), macro-F1 (ma-F1) for E-c. 
Additionally, the official evaluation also incorporates secondary evaluation metrics. For the regression tasks, they also use pearson correlation for a subset of the test set that includes only those tweets with intensity score greater or equal to 0.5. For the ordinal classification tasks, they also use pearson correlation for a subset of the test set that includes only those tweets with intensity classes low X, moderate X, or high X (where X is an emotion), and use weighted quadratic kappa on the full test set, and adopt weighted quadratic kappa on the some-emotion subset of the test set. Due to space limitations, we do not list the secondary evaluation, and its conclusions are consistent with the primary evaluation results.

For AEB-2, we apply accuracy, and macro-F1 as evaluation metrics for affective classification tasks. For regression tasks, we use the Pearson correlation coefficient (pcc) \cite{cohen2009pearson} as the evaluation metric.

\begin{table*}[]
\caption{Evaluation results on AEB-1. Some results are referenced from \protect\cite{yin2020sentibert1,zhang2020affect2,htait2021sentiment3}. 'FS' denotes few-shot method. Unmarked LLMs all adopt zero-shot method. 'ave' denotes macro-average. 'acc' denotes accuracy. 'mi-F1' denotes micro-F1. 'ma-F1' denotes macro-F1. The evaluation metric for the first four tasks is the Pearson correlation coefficient. The first line is the score of the top 1 on the SemEval-2018 Task1 leaderboard.}
\label{tab:results1}
\footnotesize
\begin{tabular}{p{2.4cm}p{0.5cm}p{0.5cm}p{0.5cm}p{0.5cm}p{0.8cm}p{0.5cm}p{0.5cm}p{0.5cm}p{0.5cm}p{0.8cm}p{0.8cm}p{0.7cm}p{0.5cm}p{0.9cm}p{0.9cm}}
\hline
\multicolumn{1}{c}{\textbf{model}} & \multicolumn{5}{c}{\textbf{EI-reg}}                                                  & \multicolumn{5}{c}{\textbf{EI-oc}}                                                   & \textbf{V-reg}   & \textbf{V-oc}    & \multicolumn{3}{c}{\textbf{E-c}}                 \\
\textbf{}                          & \textbf{ave}   & \textbf{anger} & \textbf{fear}  & \textbf{joy}   & \textbf{sadness} & \textbf{ave}   & \textbf{anger} & \textbf{fear}  & \textbf{joy}   & \textbf{sadness} & \textbf{valence} & \textbf{valence} & \textbf{acc}   & \textbf{mi-F1} & \textbf{ma-F1} \\ \hline
Leaderboard(1)                   & 0.799          & 0.827          & 0.779          & 0.792          & 0.798            & 0.695          & 0.706          & 0.637          & 0.720          & 0.717            & 0.873            & 0.856            & 0.609          & 0.724          & 0.592          \\
\hline
\textbf{}                          & \multicolumn{15}{c}{\textbf{PLMs}}                                                                                                                                                                                          \\

BERT-base                          & 0.785          & 0.800          & 0.781          & 0.783          & 0.742            & 0.683          & 0.698          & 0.656          & 0.712          & 0.665            & 0.840            & 0.805            & 0.567          & 0.718          & 0.568          \\
RoBERTa-base                       & 0.717          & 0.670          & 0.736          & 0.769          & 0.694            & 0.664          & -              & -              & -              & -                & 0.845            & 0.772            & 0.563          & 0.721          & 0.536          \\
SentiBERT                          & 0.722          & 0.724          & 0.740          & 0.731          & 0.691            & 0.665          & -              & -              & -              & -                & 0.835            & 0.763            & 0.535          & 0.700          & 0.522          \\ \hline
\textbf{}                          & \multicolumn{15}{c}{\textbf{Zero-shot/few-shot methods}}                                                                                                                                                                                                             \\
Falcon                             & 0.114          & 0.147          & 0.082          & 0.095          & 0.131            & 0.033          & 0.022          & 0.017          & 0.031          & 0.061            & 0.135            & 0.189            & 0.190          & 0.318          & 0.253          \\
Vicuna                             & 0.281          & 0.307          & 0.257          & 0.260          & 0.299            & 0.214          & 0.238          & 0.193          & 0.186          & 0.241            & 0.298            & 0.579            & 0.220          & 0.359          & 0.253          \\
LLaMA2-7B-chat                      & 0.194          & 0.176          & 0.257          & 0.097          & 0.247            & 0.120          & 0.112          & 0.138          & 0.115          & 0.114            & 0.094            & 0.497            & 0.257          & 0.414          & 0.286          \\
LLaMA2-13B-chat                     & 0.488          & 0.524          & 0.506          & 0.398          & 0.526            & 0.194          & 0.262          & 0.178          & 0.119          & 0.216            & 0.312            & 0.568            & 0.274          & 0.424          & 0.302          \\
ChatGPT                            & 0.599          & 0.637          & 0.573          & 0.569          & 0.618            & 0.455          & 0.500          & 0.428          & 0.363          & 0.529            & 0.637            & 0.748            & 0.382          & 0.546          & 0.429          \\
ChatGPT-FS                         & 0.550          & 0.572          & 0.482          & 0.587          & 0.560            & 0.473          & 0.502          & 0.410          & 0.407          & 0.573            & 0.739            & 0.791            & 0.413          & 0.563          & 0.466          \\
GPT-4                              & 0.656          & 0.699          & 0.575          & \textbf{0.686} & 0.667            & \textbf{0.620} & \textbf{0.656} & \textbf{0.579} & \textbf{0.618} & \textbf{0.629}   & 0.811            & 0.788            & 0.444          & 0.572          & 0.497          \\
GPT-4-FS                           & \textbf{0.679} & \textbf{0.704} & \textbf{0.654} & 0.679          & \textbf{0.678}   & 0.562          & 0.623          & 0.523          & 0.515          & 0.585            & \textbf{0.825}   & \textbf{0.793}   & \textbf{0.460} & \textbf{0.582} & \textbf{0.515} \\ \hline
\textbf{}                          & \multicolumn{15}{c}{\textbf{Emotion-based instruction-tuning methods}}                                                                                                                                                                                                             \\
EmoBART                            & 0.795          & 0.798          & 0.803          & 0.795          & 0.782            & 0.725          & 0.705          & 0.742          & 0.723          & 0.729            & 0.851            & 0.835            & 0.528          & 0.686          & 0.548          \\
EmoT5                              & 0.783          & 0.785          & 0.797          & 0.798          & 0.751            & 0.717          & 0.703          & 0.733          & 0.726          & 0.707            & 0.852            & 0.836            & \textbf{0.559} & \textbf{0.712} & \textbf{0.568} \\
EmoOPT                             & 0.825          & \textbf{0.827} & 0.830          & 0.837          & 0.805            & 0.753          & 0.739          & 0.751          & 0.762          & \textbf{0.759}   & \textbf{0.887}   & 0.843            & 0.532          & 0.680          & 0.550          \\
EmoBLOOM                           & 0.791          & 0.802          & 0.797          & 0.790          & 0.776            & 0.732          & 0.725          & 0.717          & 0.746          & 0.740            & 0.857            & 0.822            & 0.528          & 0.683          & 0.552          \\
EmoLLaMA-7B                        & 0.822          & 0.819          & 0.821          & 0.837          & 0.809            & 0.743          & 0.738          & 0.722          & 0.768          & 0.745            & 0.879            & 0.843            & 0.545          & 0.695          & 0.563          \\
EmoLLaMA-chat-7B                   & 0.824          & 0.825          & 0.830          & 0.832          & 0.810            & 0.751          & 0.748          & 0.754          & 0.764          & 0.739            & 0.876            & 0.827            & 0.534          & 0.693          & 0.540          \\
EmoLLaMA-chat-13B                  & \textbf{0.831} & \textbf{0.827} & \textbf{0.835} & \textbf{0.843} & \textbf{0.817}   & \textbf{0.763} & \textbf{0.755} & \textbf{0.764} & \textbf{0.777} & 0.755            & 0.886            & \textbf{0.860}   & 0.537          & 0.696          & 0.545          \\ \hline
\end{tabular}
\end{table*}

\begin{figure*}[htb]
\centering
\includegraphics[width=2\columnwidth]{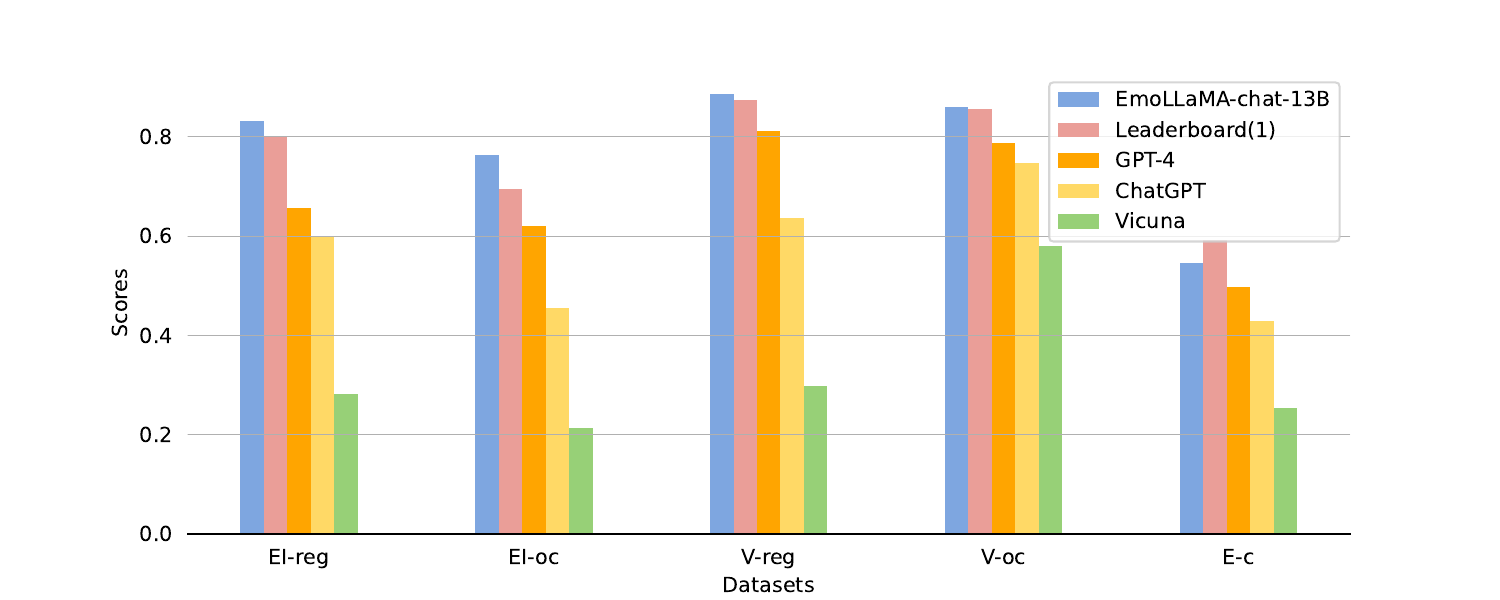}
\caption{Comparison between EmoLLMs and PLMs, Zero-shot/few-
shot methods on AEB-1. The evaluation score for the first four tasks is the pcc (EI-reg and EI-oc adopt macro-average). E-c utilizes macro-F1 score.}
\label{fig:results_AEB1}
\end{figure*}

\subsection{Results}

\subsubsection{Results on AEB-1}

The evaluation results on AEB-1 are shown in Table \ref{tab:results1} (The results of open-sourced models are the average of five runs). The first line is the score of the top 1 on the SemEval-2018 Task1 leaderboard. 

\textbf{Comparison between EmoLLMs and PLMs, Zero-shot/few-shot methods:} Figure ~\ref{fig:results_AEB1} presents the results on AEB-1 of several different kind of methods. For EmoLLMs, we chose EmoLLaMA-chat-13B, which shows the best overall performance,  to compare with other categories. The results in Table \ref{tab:results1} show that EmoLLaMA-chat-13B outperforms all other LLMs and surpasses the top ranking\footnote{Seernet \cite{duppada2018seernet} achieved first position in the first four tasks of SemEval-2018 Task1 during the competition phase. It is based on traditional machine learning methods, which perform comprehensive data pre-processing and apply the stacking technique to ensemble multiple ML methods (e.g. XG Boost, Random Forest).} in the first four tasks of AEB-1. For the complex tasks EI-reg and EI-oc, EmoLLaMA-chat-13B shows high improvement compared to top 1, with respective increases of 3.2\% (EmoLLaMA:0.831, top1:0.799) and 6.8\% (EmoLLaMA:0.763, top1:0.695). For independent raw task fine-tuning methods, although these PLMs are trained on extensive datasets and fine-tuned separately for each task, the results do not surpass the original top 1 scores. The findings demonstrate that general PLMs are more prone to overlooking important information compared to LLMs when dealing with affective regression tasks and fine-grained sentiment classification tasks. For zero-shot/few-shot methods, we can observe that this category of methods performs poorly compared to other fine-tuning approaches, especially in the EI-reg and EI-oc tasks. This indicates that the LLMs without fine-tuning struggle to handle the issue of emotion intensity effectively (We also test BART, T5, OPT and BLOOM in zero-shot and few-shot methods, but their response is highly irrelevant).

\textbf{Comparison between EmoLLMs:} We can observe from Table \ref{tab:results1} that EmoLLMs all perform well compared with LLMs without fine-tuning. EmoT5 performs the best on the emotion classification task E-c (ma-F1) (EmoT5:0.568, EmoLLaMA:0.545), but it does not perform as well as other models on regression tasks (e.g. EI-reg(ave): EmoT5:0.783, EmoLLaMA:0.831). Although EmoOPT slightly outperforms EmoLLaMA in a few regression tasks (e.g. V-reg: EmoOPT:0.887, EmoLLaMA:0.886), it still lags behind EmoLLaMA in most tasks.

In conclusion, our proposed instruction-tuning strategy for sentiment analysis tasks outperforms PLMs and all LLMs without fine-tuning, achieving the best comprehensive performance. Compared to other instruction-tuned EmoLLMs, EmoLLaMA demonstrates a more comprehensive and integrated capability in affective analysis.

\subsubsection{Results on AEB-2}

In order to evaluate the generalizability of EmoLLMs, we execute experiments on the AEB-2 that are not included in the training process (Detailed descriptions can be found in Table \ref{tab:statisticsoggtd}). All models apply zero-shot method. We compare the series of EmoLLMs with ChatGPT, GPT4, several open-source LLMs (i.e. LLaMA2-chat, Falcon, and Vicuna) and several sentiment analysis tools (i.e. VADER, TextBlob). Table \ref{tab:generaltest} presents the experiment results (The results of open-sourced models are the average of five runs). For EmoLLMs, it is worth noting that, since we use labels ranging from 0 to 1 when fine-tuning the model on the regression dataset, we also use the range of 0 to 1 for predictions during the generalization testing of regression tasks. Afterward, we map these predictions to the corresponding range of the data. 

\begin{figure*}[htb]
\centering
\includegraphics[width=2\columnwidth]{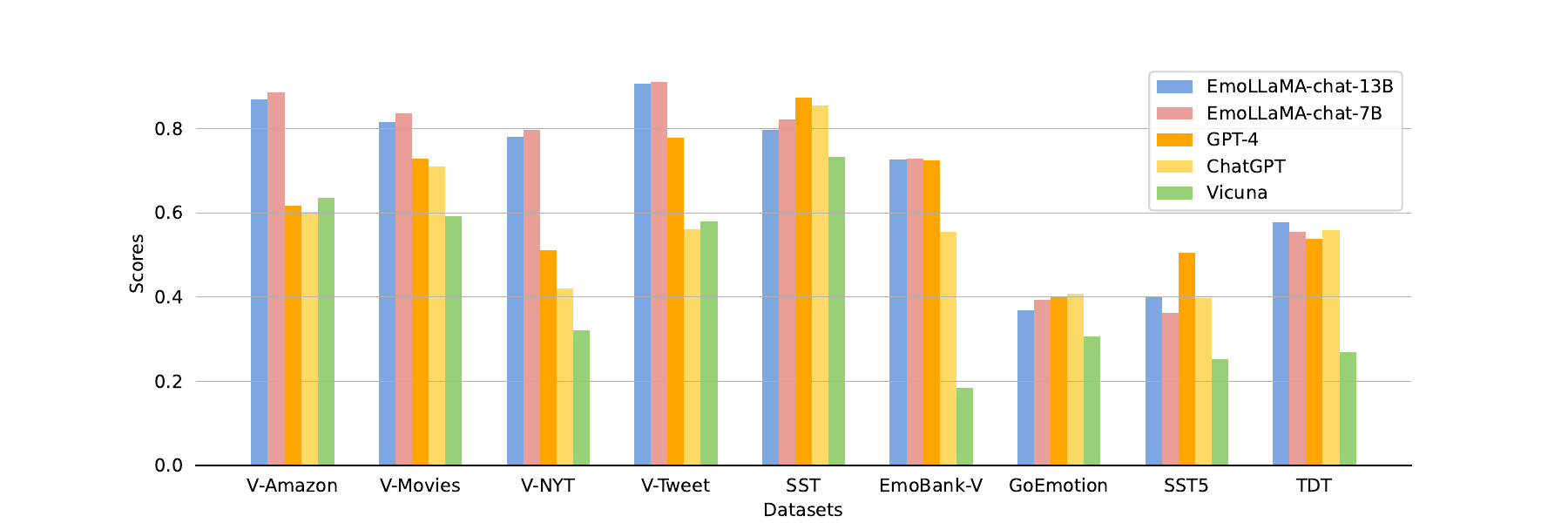}
\caption{Comparison between EmoLLMs and LLMs without fine-tuning on AEB-2. The evaluation score for the first six tasks (regression tasks) is the pcc. The last three tasks (classification tasks) utilize the macro-F1 score.}
\label{fig:results_AEB2}
\end{figure*}

\textbf{Comparision between EmoLLMs and LLMs without fine-tuning:} Figure ~\ref{fig:results_AEB2} presents the results on AEB-2 of several different kind of methods. We still choose EmoLLaMA as the representative for EmoLLMs. From Table \ref{tab:generaltest}, we can see that EmoLLaMA series outperform ChatGPT, GPT-4, and LLMs without fine-tuning in most regression tasks. In the first four regression tasks, EmoLLaMA overtakes GPT-4 by over 10\%. Although EmoLLaMA performs less well than ChatGPT and GPT-4 in SST and Emobank-Arousal, the difference is less than 5\%. For classification tasks, EmoLLaMA series performs better than ChatGPT and GPT-4 in the TDT task. In the GoEmotion, the performance of EmoLLaMA is within a 5\% difference compared to ChatGPT and GPT-4. 
In SST5 tasks, GPT-4 performs exceptionally well (acc:0.543, ma-F1:0.504), as we can see that ChatGPT, GPT-4, both outperform other models in SST5 and SST tasks. The possible reason is that the SST dataset is popular, and LLMs have been exposed to similar corpora during pre-training, which enables them to perform better using zero-shot methods. 

\textbf{Comparision between EmoLLMs:} Table ~\ref{tab:generaltest} shows that all instruction tuning LLMs perform well on AEB-2 and have good transferability except EmoBART and EmoT5. EmoBART and EmoT5 perform similarly to their performance on the AEB-1 dataset, showing poor performance in regression tasks. Interestingly, EmoLLaMA-chat-7B performs the best in most tasks of the AEB-2 and even outperforms EmoLLaMA-chat-13B in most regression tasks. One possible reason is that models with a larger number of parameters tend to overfit during fine-tuning, which can subsequently affect their general performance ability.

It is worth noting that, in AEB-2 dataset, only TDT and V-Tweet are sourced from Twitter, while the others are collected from different platforms and domains. Although EmoLLMs' training data is only sourced from Twitter, it performs well on other platforms and domains, which demonstrates its excellent transferability. The results also show that the performance of the current sentiment analysis tools (i.e. VADER, TextBlob) is significantly inferior to that of EmoLLMs. Overall, the experiment results on AEB-2 illustrate EmoLLMs series achieves ChatGPT-level and GPT-4-level general capabilities (especially EmoLLaMA) and can be used as emotion annotation tools. 

\begin{table*}[]
\centering
\caption{Evaluation results on AEB-2 dataset. All evaluation results are based on the zero-shot approach. 'pcc' denotes Pearson correlation coefficient. 'V' denotes Valence. 'V-A' denotes V-Amazon. 'V-M' denotes V-Movies. 'V-T' denotes V-Tweet. 'A' denotes arousal. 'D' denotes dominance. 'acc' denotes accuracy.}
\label{tab:generaltest}
\footnotesize
\begin{tabular}{lcccccccccccccc}
\hline
\textbf{model}    & V-A            & V-M            & V-NYT          & V-T            & SST            & \multicolumn{3}{c}{EmoBank}                      & \multicolumn{2}{c}{GoEmotion}   & \multicolumn{2}{c}{SST5}        & \multicolumn{2}{c}{TDT}         \\
                  & pcc           & pcc           & pcc           & pcc           & pcc           & V-pcc         & A-pcc         & D-pcc         & acc            & ma-F1             & acc            & ma-F1             & acc            & ma-F1             \\ \hline
VADER    & 0.565 & 0.446 & 0.464 & 0.862 & 0.450 & - & - & - & - & - & - & - & 0.510 & 0.266 \\
TextBlob & 0.490 & 0.372 & 0.288 & 0.666 & 0.408 & - & - & - & - & - & - & - & 0.434     & 0.435   \\ \hline

Falcon            & 0.492          & 0.530          & 0.340          & 0.449          & 0.205          & 0.092          & 0.059          & 0.009          & 0.168          & 0.223          & 0.231          & 0.179          & 0.355          & 0.285          \\
Vicuna            & 0.634          & 0.592          & 0.320          & 0.580          & 0.733          & 0.184          & 0.140          & 0.002          & 0.250          & 0.307          & 0.293          & 0.253          & 0.312          & 0.269          \\
LLaMA2-chat-13B    & 0.348          & 0.479          & -0.011         & 0.300          & 0.811          & 0.237          & 0.248          & -0.036         & 0.278          & 0.337          & 0.346          & 0.281          & 0.436          & 0.437          \\
ChatGPT           & 0.601          & 0.709          & 0.419          & 0.560          & 0.854          & 0.554          & 0.320          & -0.121         & 0.342          & \textbf{0.407} & 0.500          & 0.397          & 0.552          & 0.559          \\
GPT4              & 0.616          & 0.727          & 0.510          & 0.778          & \textbf{0.872} & 0.723          & \textbf{0.364} & 0.193          & 0.327          & 0.401          & \textbf{0.543} & \textbf{0.504} & 0.532          & 0.538          \\ \hline
EmoBART           & 0.770          & 0.661          & 0.650          & 0.853          & 0.634          & 0.670          & 0.059          & 0.101          & 0.318          & 0.366          & 0.341          & 0.323          & 0.529          & 0.535          \\
EmoT5             & 0.838          & 0.728          & 0.668          & 0.889          & 0.724          & 0.714          & 0.239          & 0.064          & 0.327          & 0.374          & 0.373          & 0.376          & 0.578          & \textbf{0.583} \\
EmoOPT            & 0.883          & 0.842 & 0.767          & 0.904          & 0.815          & 0.713          & 0.120          & \textbf{0.247} & 0.287          & 0.331          & 0.286          & 0.255          & 0.573          & 0.528          \\
EmoBLOOM          & 0.848          & \textbf{0.852}          & 0.649          & 0.856          & 0.807          & 0.663          & 0.106          & 0.243          & 0.302          & 0.348          & 0.409          & 0.353          & 0.546          & 0.551          \\
EmoLLaMA-7B       & 0.866          & 0.838          & 0.732          & 0.902          & 0.819          & 0.717          & 0.219          & 0.209          & 0.330          & 0.366          & 0.296          & 0.260          & \textbf{0.599} & 0.579          \\
EmoLLaMA-chat-7B  & \textbf{0.885} & 0.835          & \textbf{0.797} & \textbf{0.910} & 0.822          & \textbf{0.728} & 0.192          & 0.226          & \textbf{0.371} & 0.392          & 0.400          & 0.362          & 0.554          & 0.554          \\
EmoLLaMA-chat-13B & 0.868          & 0.815          & 0.780          & 0.906          & 0.797          & 0.726          & 0.332          & 0.218          & 0.350          & 0.369          & 0.412          & 0.399          & 0.574          & 0.578          \\ \hline
\end{tabular}
\end{table*}

\subsubsection{Analysis of Chatgpt and GPT-4 }

On the AEB-1 dataset, Table \ref{tab:results1} shows that GPT-4 and GPT-4-FS perform best in zero-shot/few-shot methods, followed by ChatGPT and ChatGPT-FS. This illustrates the current open-sourced LLMs still have a big gap with ChatGPT and GPT-4 in complex tasks (e.g. gaps between GPT-4 and Vicuna: EI-reg(ave): 0.375, EI-oc(ave): 0.426, V-reg: 0.406, V-oc: 0.513, E-c(macro-F1): 0.244). An interesting phenomenon is that in most tasks, ChatGPT and GPT-4 perform better with few-shot method than zero-shot. However, in EI-reg, ChatGPT's zero-shot method outperforms few-shot, while in EI-oc, GPT-4's zero-shot performs better than few-shot. A possible reason is that the EI-reg and EI-oc tasks are more complex, and few-shot learning requires careful design in order to improve the model's performance. For different models, there may be different understandings of few-shot examples. Therefore, for complex tasks, it is necessary to design targeted few-shot examples specifically for different LLMs.

On the AEB-2 dataset, GPT-4 and ChatGPT also perform better than other open-sourced LLMs without fine-tuning in most tasks. However, compared to the performance on the AEB-1 dataset, the performance gap between ChatGPT, GPT-4, and other LLMs without fine-tuning becomes smaller in several tasks (e.g. gaps between GPT-4 and Vicuna: V-A: -0.018, V-M: 0.135, and GoEmotion(acc):0.077). One possible reason is that these tasks are simpler compared to the tasks in AEB-1. This further demonstrates that ChatGPT and GPT-4 are more adept at handling complex tasks compared to other open-source LLMs.

To sum up, there is still a certain gap between the current open-source LLMs and ChatGPT, GPT-4 in affective analysis tasks. Currently, we can only surpass ChatGPT and GPT-4 by fine-tuning on specific tasks.

\section{Conclusion \label{sec:conclusion}}

In this paper, we propose EmoLLMs, a series of comprehensive affective analysis models and annotation tools. We also construct a multi-task affective analysis instruction dataset (AAID) and an affective evaluation benchmark (AEB). We conduct a comprehensive analysis of the performance of EmoLLMs, as well as a variety of LLMs on the AEB benchmark. The results indicate that EmoLLMs perform exceptionally well in both affective analysis regression tasks and classification tasks, achieving SOTA compared to the other open-sourced LLMs, and EmoLLMs exhibit strong transferability, as it has achieved the generalization capabilities of ChatGPT and GPT-4 in various unseen affective analysis tasks. The results also show that there is still a certain gap between the current open-sourced LLMs and ChatGPT, GPT-4 in specific domains. An ideal solution to address the issue is the instruction-tuning strategy employed in this article, which can greatly enhance the performance of LLMs in a specific domain and surpass ChatGPT and GPT-4 in most tasks.

\section{Discussions \label{sec:discussions}}

\textbf{Real-World Applications. } EmoLLMs can provide high-quality and multiple emotional information automatically, which can be used for various practical applications. For example, (1) Misinformation detection: Rumors or fake news often convey specific emotions. Affective features can help verify misinformation \cite{liu2023emotion2}. (2) Healthcare (e.g. mental health): The severity of depressive symptoms is closely related to emotions. The main reason is that individuals with depressive symptoms often struggle to regulate their emotions, leading to a decrease in emotional complexity. Therefore, emotional information is useful for diagnosing mental disorders \cite{zhang2023emotion1}. (3) Customer service (e.g. online shopping): Conducting sentiment analysis on product reviews provides valuable insights into product and service quality as well as customer experience \cite{ali2024analyzing}. 

\textbf{Limitations and Future Work. } Most of the publicly available datasets are from the internet and social media, which have different expression forms, text formats, and styles compared to other types of textual content. Thus, when applied to the real world, there may be some biases. Additionally, current EmoLLMs are limited to English text content and lack content from other languages and modalities. In the future, we will introduce more datasets from different platforms, domains, modalities, and languages into instruction-tuning data to further enhance the capabilities of EmoLLMs.

\begin{acks}
  The code in this project is based on BELLE code \cite{BELLE,ji2023exploring,wen2023chathome}. The EmoLLaMA picture in Figure \ref{fig:examplesofemollama} was generated by PIXLR\footnote{https://pixlr.com/image-generator/}. This work is supported by the computational shared facility at the University of Manchester and the scholar award from the Department of Computer Science at the University of Manchester. This work is supported by the project JPNP20006 from New Energy and Industrial Technology Development Organization (NEDO), the Centre for Digital Trust and Society at the University of Manchester, and the Manchester-Melbourne-Toronto Research Fund.
\end{acks}

\bibliographystyle{ACM-Reference-Format}
\bibliography{sample-base}

\appendix




\end{document}